\documentclass[a4paper]{elsarticle}
\usepackage{lineno}
\modulolinenumbers[1]
\usepackage[T1]{fontenc}
\usepackage{graphicx}
\usepackage{amsmath,amssymb,amsfonts}
\usepackage{float, array, multirow, color, multirow, booktabs}
\usepackage{makecell}
\usepackage{setspace}
\usepackage{hyperref}
\hypersetup{hidelinks}

\begin{document}

\begin{frontmatter}

\title{FocusNet: Classifying Better by Focusing on Confusing Classes}

\author[address1]{Xue~Zhang}
\ead{zxue2019@zju.edu.cn}

\author[address1]{Zehua~Sheng}
\ead{shengzehua@zju.edu.cn}

\author[address1]{Hui-Liang~Shen\corref{correspondingauthor}}
\cortext[correspondingauthor]{Corresponding author}
\ead{shenhl@zju.edu.cn}

\address[address1]{College of Information Science and Electronic Engineering, Zhejiang University, Hangzhou 310027, China}

\begin{abstract}
Nowadays, most classification networks use one-hot encoding to represent categorical data because of its simplicity. However, one-hot encoding may affect the generalization ability as it neglects inter-class correlations. We observe that, even when a neural network trained with one-hot labels produces incorrect predictions, it still pays attention to the target image region and reveals which classes confuse the network. Inspired by this observation, we propose a confusion-focusing mechanism to address the class-confusion issue. Our confusion-focusing mechanism is implemented by a two-branch network architecture. Its baseline branch generates confusing classes, and its FocusNet branch, whose architecture is flexible, discriminates correct labels from these confusing classes. We also introduce a novel focus-picking loss function to improve classification accuracy by encouraging FocusNet to focus on the most confusing classes. The experimental results validate that our FocusNet is effective for image classification on common datasets, and that our focus-picking loss function can also benefit the current neural networks in improving their classification accuracy. Models and code are available at {\color{magenta} \url{https://github.com/XueZ-phd/FocusNet-Classifying-better-by-focusing-on-confusing-classes}}.

\end{abstract}

\begin{keyword}
	Image classification, inter-class correlations, confusing classes.
\end{keyword}

\end{frontmatter}

%\linenumbers

%%%%%%%%%%%%%%%%%%%%%%%%%%%%%%%%%%%%%%%%%%%%%%%%%%%%%%%%%%%%%%%%%%%%%%%%%%%%%%%%%%%%%%%%%%%%%%%%%%%%%%%%%%%%%%%%%%%%%%%%%%%%%%%%%%%%%%%%%%%%%%%%%%%%%%%%%%%%%%%%%%%%%%%%%%%%%%%%%%%%%%%%%%%%%%%%%%%%%%%%%%%%%%%%%%%%%%%%%%
\section{Introduction}\label{sec:introduction}

\begin{figure*}[t]
\centering
\includegraphics[width=\linewidth]{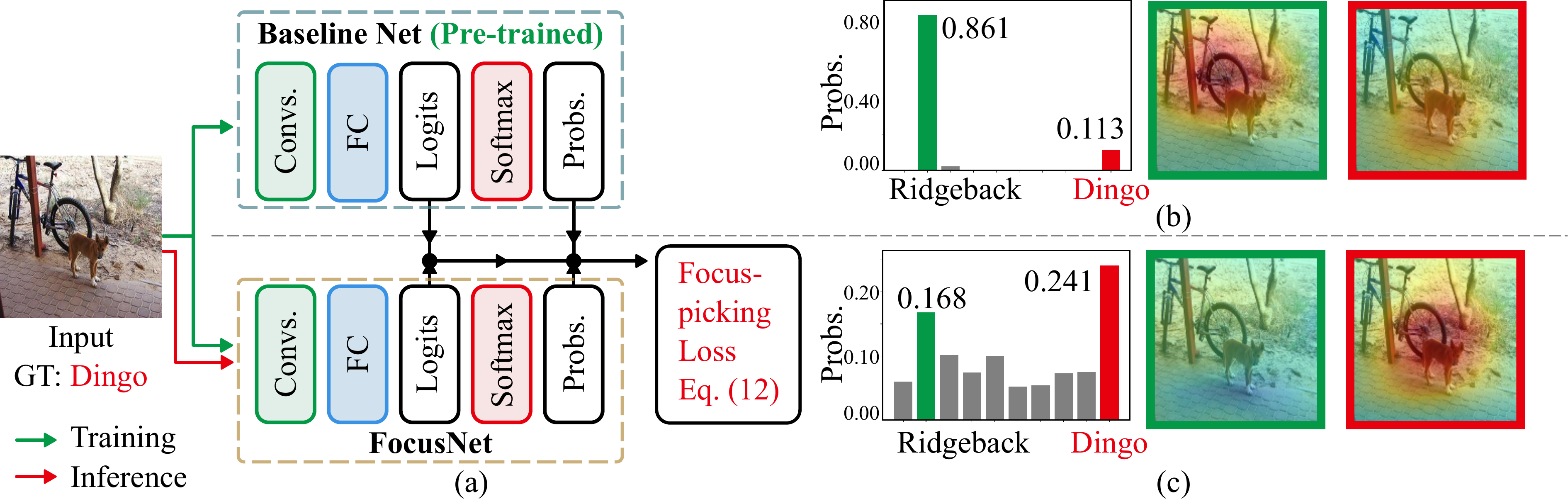}
\caption{The overall architecture of our confusion-focusing mechanism. (a) Training and inference framework of our approach. After training, only the FocusNet branch is needed for inference. (b) Prediction of the baseline network. (c) Predictions of our FocusNet. The pictures next to the bar charts are Ridgeback's CAM (green border) and Dingo's CAM (red border). Note that the baseline and FocusNet can use the same or different network architectures. Here, for demonstration, the network architectures are both ResNet-18~\cite{he2016deep}, and the dataset is Imagewoof~\cite{imagewoof}.}
\label{fig:ScoreAndCam}
\end{figure*}

Neural network based image classification has received much attention in recent years due to the impressive performance of deep learning~\cite{he2016deep, sandler2018mobilenetv2, hinton2015distilling}. Most classification networks use one-hot encoding to represent categorical data. However, this encoding method neglects inter-class correlations, and may cause over-fitting as it tends to make the network produce overconfident predictions on the training sets~\cite{szegedy2016rethinking, Mller2019WhenDL}. To address this problem, label smoothing regularization~\cite{szegedy2016rethinking} introduces a uniform distribution to encode correlations between classes, which encourages networks to be less confident. Considering the uniform distribution is independent of the training data, some approaches~\cite{hinton2015distilling, yuan2020revisiting} use soft targets to replace it with a learned distribution. Alternatively, DropMax~\cite{lee2018dropmax} drops non-target classes to select similar ones that confuse networks by using an additional multilayer perceptron. To our knowledge, there are few works exploring the confusing-class correlation for the purpose of image classification. 
	
In this work, we propose the confusion-focusing mechanism and a focus-picking loss function to handle the class-confusion issue. Fig.~\ref{fig:ScoreAndCam} illustrates the overall architecture of our confusion-focusing mechanism, which consists of a baseline branch and a FocusNet branch. The baseline network is pre-trained to produce reference predictions. FocusNet, which can be implemented using common models such as ResNet~\cite{he2016deep} and MobileNet~\cite{sandler2018mobilenetv2}, identifies confusing classes and difficult samples based on the output of the baseline network. Furthermore, our proposed focus-picking loss function focuses more on these confusing classes and assigns larger losses to difficult samples in the training stage. It is worth noting that, only the FocusNet branch is needed for inference, so our approach does not add extra computing complexity. With this strategy, FocusNet can effectively improve the performance of the baseline network. As illustrated in Fig.~\ref{fig:ScoreAndCam}~(b) and (c), compared with the baseline network, FocusNet pays more attention to the target region and produces the correct prediction. To summarize, the main contributions of this work are threefold:
\begin{itemize}
	\item We propose the confusion-focusing mechanism implemented by a two-branch network architecture to address the challenging class-confusion issue. The FocusNet branch can improve the performance by focusing on confusing classes derived from the baseline branch without adding extra computing complexity during inference.
	
	\item We introduce a focus-picking loss function for network training. Compared with normal cross-entropy, our loss function enables the network to focus on confusing classes and assign larger losses to difficult samples.
	
	\item We experimentally validate on various datasets that FocusNet outperforms the existing approaches in addressing the class-confusion issue, and that our focus-picking loss function can benefit the current knowledge distillation method in the accuracy improvement.
	
\end{itemize}

%%%%%%%%%%%%%%%%%%%%%%%%%%%%%%%%%%%%%%%%%%%%%%%%%%%%%%%%%%%%%%%%%%%%%%%%%%%%%%%%%%%%%%%%%%%%%%%%%%%%%%%%%%%%%%%%%%%%%%%%%%%%%%%%%%%%%%%%%%%%%%%%%%%%%%%%%%%%%%%%%%%%%%%%%%%%%%%%%%%%%%%%%%%%%%%%%%%%%%%%%%%%%%%%%%%%%%%%%%

\section{Related Work}\label{sec:realtedWork}

We briefly review the related works of network architecture design and loss function design, which are relevant to our novel network mechanism and loss function proposed in this work.

\subsection{Network Architecture Design}
Over the past decade, many advanced network architectures have been introduced to improve accuracy in image classification, including networks that increase in depth~\cite{he2016deep}, width~\cite{szegedy2016rethinking}, cardinality~\cite{xie2017aggregated} and scale~\cite{gao2019res2net}. Recently, inter-class correlations~\cite{szegedy2016rethinking, hinton2015distilling} and attention mechanisms~\cite{obeso2022visual, CHEN2022108567}, as two approaches to further improve the performance of existing networks, have attracted much attention.

It has been found that learning inter-class correlations is an effective way to improve the generalization ability because it prevents the networks from producing overconfident predictions~\cite{szegedy2016rethinking}. There are two main ways to use the correlation between classes. The first way uses the manually designed distribution over classes, which does not change the original network architecture. For example, label smoothing regularization~\cite{szegedy2016rethinking} mixes a uniform distribution with the ground-truth distribution by a smoothing factor. The work~\cite{pereyra2017regularizing} introduces a random relationship between classes by encouraging networks to produce high entropy output distribution. The second way uses the learned distribution for each sample and typically designs a network architecture. For example, the label embedding network~\cite{LIU2022108584} generates a soft distribution to represent the continuous interactions between classes. RDCN~\cite{liu2021relation} introduces a relation network to measure the relationship between the visual features and embedded semantics. Knowledge distillation~\cite{hinton2015distilling} introduces soft targets to replace the hand-designed distribution. Standard knowledge distillation~\cite{hinton2015distilling, shi2022explainable} is a combination of a pre-trained teacher model, and a smaller student model to be deployed. The recent self-knowledge distillation mechanism~\cite{zhang2019your, ji2021refine} refines knowledge using an auxiliary self-teacher model. Teach-free knowledge distillation~\cite{yuan2020revisiting} combines the knowledge distillation and the label smoothing regularization to produce a more accurate distribution over classes. Instead of considering the correlation between all classes, DropMax~\cite{lee2018dropmax} learns the class retain probability of each sample to drop non-target classes and select the confusing ones. 

Another effective approach to improve the accuracy of networks is to use the attention mechanism. For example, VINet~\cite{obeso2022visual} integrates human visual attention into neural networks in image classification and object detection. AE-Net~\cite{CHEN2022108291} refines channel and spatial attention to retrieve fine-grained sketch-based images. DAAF~\cite{CHEN2022108567} learns global and local attention aware features to improve the accuracy of person re-identification. Progressive-attention network~\cite{zheng2019learning} localizes discriminative parts at multiple scales progressively. 

The main difference between these methods and our work is that FocusNet establishes the relationship between confusing classes and network attention. In addition, our FocusNet does not change the architecture of the original network, and does not add extra computing complexity during inference.

\subsection{Loss Function Design}
In supervised learning, networks are trained through an optimization process that minimizes a loss function. In this case, most networks for multi-class classification use the softmax cross-entropy as their loss function~\cite{he2016deep}. Minimizing the cross-entropy is equivalent to minimizing the Kullback-Leibler divergence (hereafter written as KL divergence) when measuring the degree of dissimilarity between empirical distribution and model distribution~\cite{CHEN2021107983}. Currently, some variants of the softmax cross-entropy have been introduced to solve task-specific issues~\cite{hinton2015distilling, lee2018dropmax, lin2017focal}. Focal loss~\cite{lin2017focal} addresses the class imbalance problem by introducing a tunable weighting factor to the cross-entropy. Adjusting this weighting factor allows networks to reduce relative losses for well-classified samples and focus more on hard and misclassified ones. DropMax~\cite{lee2018dropmax} focuses on confusing classes by dropping non-target logits in the softmax layer, which uses the Bayesian inference to calculate class retain probabilities and uses variational inference to optimize parameters. Knowledge distillation~\cite{hinton2015distilling} transfers knowledge from a large teacher to an easy-deployed student by minimizing the KL divergence of soft targets between the teacher and the student, which is mathematically equivalent to optimizing the mean square error of logits between the two models. LSA~\cite{liu2022zero} considers the low-rank structure of the reconstruction data in tackling the domain bias problem.

Unlike the above methods, our focus-picking loss function mainly considers the most confusing inter-class interactions and assigns larger losses to more difficult samples.

%%%%%%%%%%%%%%%%%%%%%%%%%%%%%%%%%%%%%%%%%%%%%%%%%%%%%%%%%%%%%%%%%%%%%%%%%%%%%%%%%%%%%%%%%%%%%%%%%%%%%%%%%%%%%%%%%%%%%%%%%%%%%%%%%%%%%%%%%%%%%%%%%%%%%%%%%%%%%%%%%%%%%%%%%%%%%%%%%%%%%%%%%%%%%%%%%%%%%%%%%%%%%%%%%%%%%%%%%%
\section{Proposed Method}\label{sec:proposedAlg}
In this section, we first introduce our motivation. Then, we elaborate on the proposed FocusNet and the focus-picking loss function. To better clarify the concept of focus-picking loss, we also analyze its mathematical mechanism.

\subsection{Motivation}\label{sec:motivation}
We notice that similar classes tend to confuse networks. As illustrated in Fig.~\ref{fig:ScoreAndCam}~(b), the baseline network yields an incorrect prediction. The ground-truth class is Dingo, but the top-1 predicted one is Ridgeback.  Nevertheless, the CAM of the class Dingo (red border) shows that the baseline network has paid attention to the target image region, \emph{i.e.} the Dingo's body. In other words, the baseline network has noticed the correct class, but it is confused by incorrect classes and cannot make the correct prediction. Inspired by this, we aim to address the class-confusion issue by making networks discriminate the correct class from these confusing ones.

Fig.~\ref{fig:ScoreAndCam}~(a) illustrates our proposed confusion-focusing mechanism, which combines a pre-trained baseline network branch (for generating reference distributions at training time) and a FocusNet branch (for making predictions). It is worth noting that the FocusNet branch is not restricted to a specific network architecture. It uses the reference distribution derived from the baseline network to determine confusing classes based on the above observation. In addition, FocusNet compares the probability on the correct class with the reference distribution to identify difficult samples. In this context, we design the focus-picking loss function to focus on picking the correct class from the confusing ones during training. The baseline network branch is removed during inference to keep the model size of our approach unchanged. With this approach, Fig.~\ref{fig:ScoreAndCam}~(c) shows that our FocusNet correctly predicts Dingo. Besides, the CAM of the class Dingo (red border) illustrates that our FocusNet pays more attention to the target image region.

From the perspective of requiring prior knowledge from a pre-trained network, the network architecture of our approach is similar to standard knowledge distillation to some extent. But we have validated that our confusion-focusing mechanism is effective even when 1) the pre-trained baseline network uses various network scales, \emph{i.e.}, the same as, or larger or smaller than FocusNet architecture, and 2) the pre-trained network is highly incorrect. In comparison, however, the standard knowledge distillation generally requires a larger and more accurate pre-trained model. Please refer to Section~\ref{sec:diffKD} for more detailed analysis.

\subsection{Confusion-focusing Mechanism}
\label{sec:FocusNetMechanism}
As mentioned, our confusion-focusing mechanism can obtain consistent performance improvements with a variety of network architecture combinations. Fig.~\ref{fig:ScoreAndCam}~(a) illustrates an example pipeline that adopts ResNet-18~\cite{he2016deep} as the network architecture of the baseline network branch and the FocusNet branch.

\subsubsection{Training and Inference of the Baseline Network Branch}
Given an input, the baseline network branch in Fig.~\ref{fig:ScoreAndCam}~(a) produces logits $\mathbf{z}^b$ through a sequence of convolutional layers and a fully-connected layer. Then the resulting logits are converted to the probability distribution $\mathbf{\hat{y}}^b$ using a softmax function. The process can be formulated as
\begin{eqnarray}\label{eq:baselineLogits}
\mathbf{z}^b = \left(\mathbf{W}^b\right)^{\mathsf{T}}\mathbf{h}^b,
\end{eqnarray}
\begin{eqnarray}\label{eq:baselineSoftmax}
	\hat{y}^b_n = \frac{\exp(z^b_n)}{\sum_{j=1}^{N}{\exp(z^b_j)}},
\end{eqnarray}
where matrix $\mathbf{W}^b=\left(\mathbf{w}^b_1, \mathbf{w}^b_2, \cdots, \mathbf{w}^b_N\right)$ contains the weights and biases of the fully-connected layer in the baseline network branch, and the column vector $\mathbf{w}^b_n, 1\leq n \leq N,$ denotes the template of the \textit{n}-th class~\cite{Mller2019WhenDL}. Vector $\mathbf{h}^b$ contains the activations of the penultimate layer concatenated with ``1'' accounting for the bias. $ z_n^b $ and $\hat{y}^b_n$ denote the $n$-th entries of the vectors $\mathbf{z}^b$ and $\mathbf{\hat{y}}^b$, respectively.

The baseline network branch is trained by computing the cross-entropy between the one-hot label $\mathbf{y}$ and probabilities $\mathbf{\hat{y}}^b$,
\begin{eqnarray}\label{eq:baselineCE}
	\begin{aligned}
		\mathcal{L}^{b} ={}&H(\mathbf{y}, \mathbf{\hat{y}}^b)
		\\={}& -\sum_{n=1}^{N}y_n\log(\hat{y}^b_n),	
	\end{aligned}
\end{eqnarray}
where $ y_n $ denotes the $n$-th entry of $\mathbf{y}$.

\subsubsection{Training and Inference of the FocusNet Branch}
Our FocusNet is trained on the same dataset by minimizing the proposed focus-picking loss function. Specifically, the green arrows in Fig.~\ref{fig:ScoreAndCam}~(a) illustrate that the training framework of our approach consists of a pre-trained baseline network branch and a FocusNet branch. We note that the baseline branch is fixed when training our FocusNet. Each training sample is fed into both branches at the same time to produce two sets of logits and probabilities. Then, FocusNet uses the logits of the baseline branch to deduce the multi-warm label that represents the confusing classes. In addition, FocusNet compares the probabilities on the ground-truth class with the baseline prediction to determine if the training sample is difficult to classify. Our focus-picking loss function encourages FocusNet to learn the multi-warm label and adaptively tunes its value according to the classification difficulty of the training sample.

After completing the training stage, we only need the FocusNet branch for inference, as the red arrow in Fig.~\ref{fig:ScoreAndCam}~(a) indicates. As a result, our approach adds no extra computing complexity during inference.

%%%%%%%%%%%%%%%%%%%%%%%%%%%%%%%%%%%%%%%%%%%%%%%%%%%%%%%%%%%%%%%%%%%%%%%%%%%%%%%%%%%%%%%%%%%%%%%%%%%%%%%%%%%%%%%%%%%%%%%%%%%%%%%%%%%%%%%%%%%%%%%%%
\subsection{Multi-warm Label}
\label{sec:mlLabel}
To guide the computation of FocusNet attention, we employ the pre-trained baseline network branch to determine confusing classes for each training sample, mathematically represented as a multi-warm label. Specifically, the baseline network generates logits over classes based on Eq.~\eqref{eq:baselineLogits}. In this context, the logit $ z_n^b=\left(\mathbf{w}^b_n\right)^\mathsf{T}\mathbf{h}^b $ of the \textit{n}-th class measures the correlation between the vector $\mathbf{h}^b$ and the template $\mathbf{w}^b_n$. A positive or negative correlation value indicates that the vector and the template point in close or opposite directions, while a zero value indicates that they are orthogonal.

The \textit{n}-th entry of the multi-warm label vector is defined as
\begin{eqnarray}\label{eq:ml1}
	l'_n = \left\{
	\begin{array}{rl}
		1, & \text{if }z_n^b>0\\
		0, & \text{if }z_n^b\leq0.
	\end{array}\right.
\end{eqnarray}
Then, we add the ground-truth label of the training sample to the multi-warm label and clip it for learning stability~\cite{lee2018dropmax},
\begin{eqnarray}\label{eq:ml2}
	l''_n = \min\left(1, l'_n+y_n\right).
\end{eqnarray}
After normalizing each label entry by
\begin{eqnarray}\label{eq:multi-warmLabel}
	l_n = \frac{l''_n}{\sum_{j=1}^N{l''_j}},
\end{eqnarray}
we obtain the final multi-warm label $\mathbf{l}=(l_1,l_2,\cdots,l_{N})^\mathsf{T}$. The non-zero entries of the label vector $\mathbf{l}$ have the same probability and sum to 1.0.

\begin{figure*}[t]
	\centering
	\includegraphics[width=\linewidth]{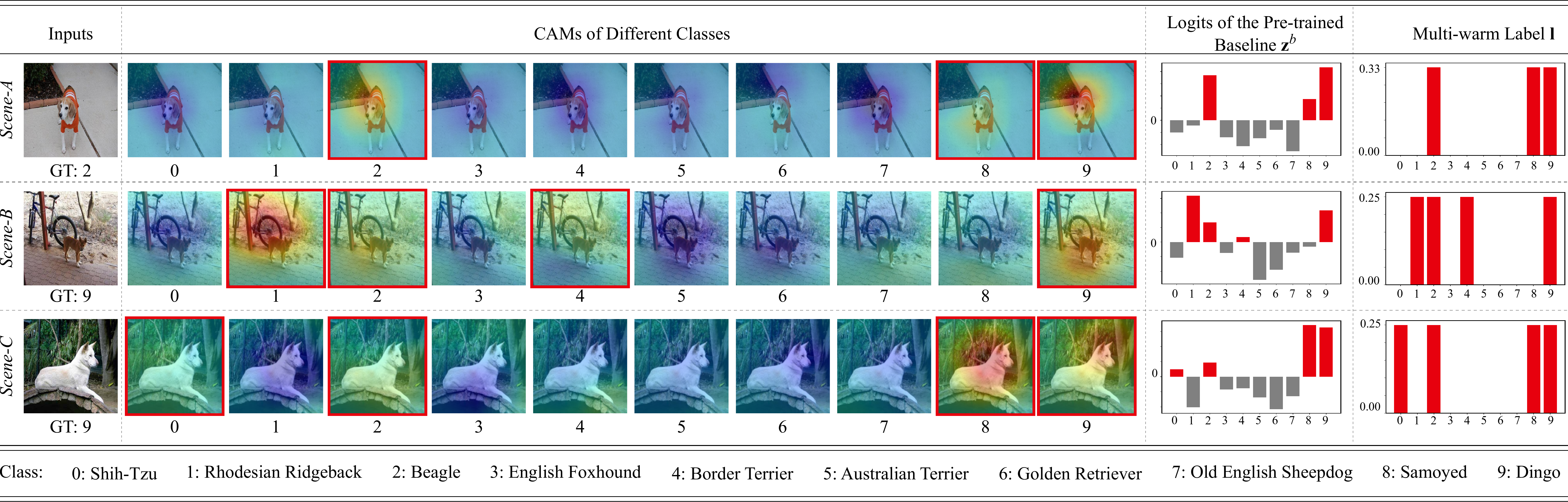}
	\caption{The relationship between the attention of the pre-trained baseline network and our proposed multi-warm label. The pre-trained baseline network processes the input image in each scene to generate CAMs, logits, and the multi-warm label. CAMs highlight discriminative image regions for corresponding classes. The positive and negative logits are marked in red and gray, respectively. Non-zero entries of the multi-warm label and the CAMs of confusing classes are marked in red bars and red borders, respectively. The ``Class'' row presents categorical indexes and categorical names.}
	\label{fig:logitsAndCAMs}
\end{figure*}
%%%%%%%%%%%%%%%%%%%%%%%%%%%%%%%%%%%%%%%%%%%%%%%%%%%%%%%%%%%%%%%%%%%%%%%%%%%%%%%%%%%%%%%%%%%%%%%%%%%%%%%%%%%%%%%%%%%%%%%%%%%%%%%%%%%%%%%%%%%%%%%%%%%%%%%%%%%%%%%%%%%%%%%%%%%%%%%%%%%%%%%%%%%%%%%%%%%%%%%

The multi-warm label reflects the attention of the pre-trained baseline network. In other words, it converts the implicit attention of the baseline network into explicit knowledge that FocusNet can learn directly. In Fig.~\ref{fig:logitsAndCAMs}, we show the prediction examples of three scenes, each of which contains the CAMs of different classes, the corresponding logits, and the multi-warm labels. For \emph{Scene-A}, the baseline network pays attention to Beagle (index: 2, ground truth), Samoyed (index: 8), and Dingo (index: 9). It generates positive logits on these classes as they are similar in appearance, implying that the multi-warm label reflects the attention of the baseline network. \emph{Scene-B} and \emph{Scene-C} both contain Dingoes, but with different fur colors (yellow and white, respectively). The baseline network generates significantly different CAMs and multi-warm labels, demonstrating that the multi-warm labels are adaptive. In this context, the multi-warm label can indicate which classes are confusing.
%%%%%%%%%%%%%%%%%%%%%%%%%%%%%%%%%%%%%%%%%%%%%%%%%%%%%%%%%%%%%%%%%%%%%%%%%%%%%%%%%%%%%%%%%%%%%%%%%%%%%%%%%%%%%%%%%%%%%%%%%%%%%%%%%%%%%

\subsection{Loss Function}\label{sec:loss}
FocusNet converts logits $\mathbf{z}$ into predicted probabilities $\mathbf{\hat{y}}$, computed by
\begin{eqnarray}\label{eq:clonalSoftmax}
	\hat{y}_n = \frac{\exp{\left({d}_n+{z}_n\right)}}{\sum_{j=1}^{N}{\exp{\left({d}_j+{z}_j\right)}}}.
\end{eqnarray}
Here,
\begin{eqnarray}\label{eq:difference}
	d_n = \frac{\exp(z_n)}{\sum_{j=1}^{N}{\exp(z_j)}}-\frac{\exp(z^b_n)}{\sum_{j=1}^{N}{\exp(z^b_j)}},
\end{eqnarray}
is the $ n $-th entry of the vector $\mathbf{d}$. $ z_n $ and $ \hat{y}_n $ are the logit and the predicted probability of FocusNet on the $ n $-th class, respectively. Vector $\mathbf{d}$ denotes the probability difference over classes between FocusNet and the pre-trained baseline network. The normal softmax probability distribution in Eq.~\eqref{eq:baselineSoftmax} is a special case of the re-weighted probability distribution in Eq.~\eqref{eq:clonalSoftmax} when $\mathbf{d}=\mathbf{0}$. For notation simplicity, we denote the normal and re-weighted probability distributions of FocusNet as $\mathbf{\hat{y}}(\mathbf{z}, \mathbf{d}=\mathbf{0})$ and $\mathbf{\hat{y}}(\mathbf{z}, \mathbf{d})$, respectively. A positive difference on the ground-truth class means that the input is a relatively easy sample since FocusNet generates a more confident prediction than the baseline network. In contrast, a negative difference corresponds to a relatively difficult sample. Here, the role of difference is to force FocusNet to tune the weights for logits so that easy samples and difficult samples can produce higher and lower probabilities on the ground-truth class, respectively.

Then, we compute the cross-entropy between the re-weighted probabilities and the ground-truth label as the classification loss
\begin{eqnarray}\label{eq:re-weightedCE}
	\begin{aligned}
		\mathcal{L}_{\rm cls} ={}& H\left(\mathbf{y}, \mathbf{\hat{y}}(\mathbf{z}, \mathbf{d})\right)\\ ={}&-\sum_{n=1}^Ny_n\log\left(\hat{y}_n(\mathbf{z}, \mathbf{d})\right)\\
		={}&-\log\left(\frac{\exp(d_t+z_t)}{\sum_{j=1}^N\exp(d_j+z_j)}\right),
	\end{aligned}
\end{eqnarray}
where index $t$ refers to the ground-truth class. This allows FocusNet to take the prediction from the pre-trained baseline network as a reference and assigns larger losses to those relatively difficult samples (see Section~\ref{sec:FPLossMechanism} for more details). At test time, FocusNet computes the prediction by setting $\mathbf{d}=\mathbf{0}$, \emph{i.e.} the normal softmax function.

To make FocusNet pay more attention to confusing classes during training, we propose to regularize it by minimizing the cross-entropy between normal softmax probabilities and the multi-warm label. That is,
\begin{eqnarray}\label{eq:attentionR}
	\begin{aligned}
		\mathcal{R}_{\rm attention} ={}&H\left(\mathbf{l}, \mathbf{\hat{y}}(\mathbf{z}, \mathbf{d}=\mathbf{0})\right)\\
		={}&-\sum_{n=1}^Nl_n\log\left(\hat{y}_n(\mathbf{z}, \mathbf{d}=\mathbf{0})\right)\\
		={}&-\sum_{n=1}^Nl_n\log\left(\frac{\exp(z_n)}{\sum_{j=1}^N\exp(z_j)}\right).
	\end{aligned}
\end{eqnarray}
This regularization encourages FocusNet to produce higher probabilities on confusing classes than the others. For example, \textit{Scene-B} in Fig.~\ref{fig:logitsAndCAMs} shows that the baseline network is confused by classes 1, 2, 4, and 9, respectively. Using our approach in Eq.~\eqref{eq:attentionR}, FocusNet generates higher probabilities on these four classes than the others, as shown in the FocusNet branch of Fig.~\ref{fig:ScoreAndCam}~(c).

Last but not least, minimizing the classification loss in Eq.~\eqref{eq:re-weightedCE} encourages FocusNet to make more confident predictions on ground-truth classes than the baseline network. This may produce overconfident distributions, \emph{i.e.} low entropy distributions~\cite{pereyra2017regularizing}. To reduce this potential degradation, we penalize the low-entropy distributions computed as
\begin{eqnarray}\label{eq:entropyR}
	\begin{aligned}
		\mathcal{R}_{\rm entropy} ={}&H\left(\mathbf{\hat{y}}(\mathbf{z}, \mathbf{d}=\mathbf{0})\right)\\
		={}&-\sum_{n=1}^N\hat{y}_n(\mathbf{z}, \mathbf{d}=\mathbf{0})\log\left(\hat{y}_n(\mathbf{z}, \mathbf{d}=\mathbf{0})\right)\\
		={}&-\sum_{n=1}^N\frac{\exp(z_n)}{\sum_{j=1}^N\exp(z_j)}\log\left(\frac{\exp(z_n)}{\sum_{j=1}^N\exp(z_j)}\right).
	\end{aligned}
\end{eqnarray}
Adding negative entropy to the classification loss allows FocusNet to produce smoother probability distributions with higher entropy and thus contributes to better generalization, which is similar to label smoothing regularization~\cite{szegedy2016rethinking}.

To train FocusNet, we propose the focus-picking loss function, which consists of the three components described above. It is defined as
\begin{eqnarray}\label{eq:clonalCE}
	\begin{aligned}
		\mathcal{L} ={}& \mathcal{L}_{\rm cls}+\alpha\mathcal{R}_{\rm attention}-\beta\mathcal{R}_{\rm entropy}\\
		={}&H\left(\mathbf{y}, \mathbf{\hat{y}}(\mathbf{z}, \mathbf{d})\right)+\alpha H\left(\mathbf{l},\mathbf{\hat{y}}(\mathbf{z}, \mathbf{d}=\mathbf{0})\right)-\beta H\left(\mathbf{\hat{y}}(\mathbf{z}, \mathbf{d}=\mathbf{0})\right).
	\end{aligned}
\end{eqnarray}
It encourages FocusNet to generate larger losses on relatively difficult samples, focus on confusing classes, and prevent the low-entropy output distributions at the same time. We experimentally find that setting $\alpha=\beta=1$ works well.

\subsection{Analysis of the Focus-Picking Loss Function}
\label{sec:FPLossMechanism}
Different from traditional loss functions, our focus-picking loss function considers confusing inter-class correlations and assigns larger loss values to those difficult samples. For instance, the widely used cross-entropy with one-hot labels does not consider the interaction between classes~\cite{szegedy2016rethinking} and the difficulty of each sample. KL divergence with soft targets is usually used in knowledge distillation~\cite{hinton2015distilling, yuan2020revisiting}. But the soft targets tend to be a uniform distribution when the temperature is high~\cite{BISWAS2022108301}, so it can not emphasize the importance of confusing classes. The focal loss~\cite{lin2017focal} is a dynamically scaled cross-entropy to address the class imbalance problem in object detection, but it also does not consider the class-confusion issue.
	
\begin{figure}[t]
\centering
\includegraphics[width=0.7\linewidth]{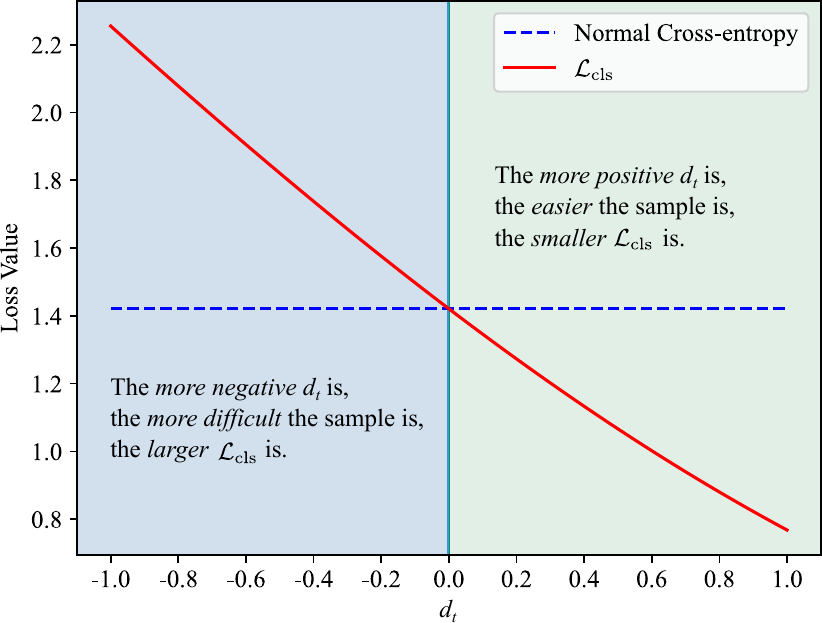}
\caption{The impact of $d_t$ on $\mathcal{L}_\mathrm{cls}$. $d_t$ denotes the probability difference on the correct class between FocusNet and the baseline network. The two descriptions in the illustration present our motivation for designing $\mathcal{L}_\mathrm{cls}$.}
\label{fig:LclsvsCE}
\end{figure}	

We first explain the data term $\cal{L}_\mathrm{cls} $. It takes the probability distribution derived from the baseline network as a reference, down-weights the contribution of easy samples during training, and rapidly enables FocusNet to focus on difficult samples. We rewrite Eq.~\eqref{eq:re-weightedCE} as
\begin{eqnarray}\label{eq:variantLcls}
	\begin{aligned}
		\mathcal{L}_{\rm cls} ={}&-\log\left(\frac{\exp(d_t+z_t)}{\sum_{j=1}^N\exp(d_j+z_j)}\right)\\
		={}&-\log\left(\frac{\exp(z_t)}{\exp(z_t)+\sum_{j\neq t}^{N}\frac{\exp(d_j)}{\exp(d_t)}\exp(z_j)}\right),
	\end{aligned}
\end{eqnarray}
where $d_t = \hat{y}_t(\mathbf{z}, \mathbf{d}=\mathbf{0})-\hat{y}_t^b$  denotes the probability difference on the ground-truth class between FocusNet and the baseline network.

Fig.~\ref{fig:LclsvsCE} shows that $\cal{L}_\mathrm{cls} $ becomes progressively smaller as $d_t $ changes from negative to positive. Specifically, a negative $d_t $ indicates that FocusNet has lower confidence on the correct class than the baseline network, which implies the sample is difficult, and thus $\cal{L}_\mathrm{cls} $ is large. A positive $d_t $ means the sample is easy, and hence $\cal{L}_\mathrm{cls} $ is small.
	
$\mathcal{R}_\mathrm{attention} $ encourages FocusNet to focus on the classes that confuse the baseline network. One can compute the derivative of $\mathcal{R}_\mathrm{attention} $ with respect to logits $\mathbf{z} $ as
\begin{eqnarray}\label{eq:RattGradient}
	\dfrac{\partial{\mathcal{R}_\mathrm{attention}}}{\partial \mathbf{z}}=\mathbf{\hat{y}(z, d=0)-\mathbf{l}},
\end{eqnarray}
where $\mathbf{l} $ is the multi-warm label encoding confusing classes, $\mathbf{\hat{y}(z, d=0)} $ is the predicted probability distribution of FocusNet. Making the derivative equal to $\mathbf{0} $ enables FocusNet to generate higher probabilities over the confusing classes while penalizing the outputs of irrelevant classes.

$\mathcal{R}_\mathrm{entropy} $ prevents FocusNet from producing over-confident predictions~\cite{pereyra2017regularizing}. It is necessary because $\mathcal{L}_\mathrm{cls} $ encourages FocusNet to make more confident predictions on correct classes than the baseline network, as discussed in Section~\ref{sec:loss}.

\subsection{Differences from Standard Knowledge Distillation}
\label{sec:diffKD}
Our proposed confusion-focusing mechanism differs from standard knowledge distillation mainly in the following four aspects:
\begin{enumerate}[1)]
	\item	Motivations are different. As described in Section~\ref{sec:motivation}, our motivation is to make FocusNet discriminate the correct class from the most confusing ones. But standard knowledge distillation aims to transfer knowledge from a teacher (a larger model or an ensemble of models) to a student (an easy-to-deployed model). We have validated in Table~\ref{tb:vsReverseKD} that a poorly pre-trained network has less influence on our FocusNet than on the knowledge distillation.
	\item	In the case of learning confusing classes, the network architectures are different. FocusNet only trains one model. It directly makes predictions without using the pre-trained baseline network at test time. However, in the case of learning confusing classes, standard knowledge distillation needs to train a set of specialist models. It first runs a pre-trained generalist model to decide which specialist models are relevant at test time. Then these specialist models are run to determine the final prediction (please refer to Section 5 and Section 7 in standard knowledge distillation~\cite{hinton2015distilling} for more details). Therefore, our proposed FocusNet is more efficient for learning confusing classes.
	\item	Labels are different. Our proposed multi-warm label emphasizes the importance of confusing classes. Soft targets~\cite{hinton2015distilling} of standard knowledge distillation considers connections between all classes, but they are difficult to reflect the relationship between confusing classes. A convincing example is that as the temperature increases, the soft target tends to be a uniform distribution~\cite{BISWAS2022108301}, which means that the correct class has a close probability of being similar to all other classes.
	\item	Loss functions are different. The proposed focus-picking loss function enables FocusNet to focus on confusing classes and assign larger losses to more difficult samples (as described in Section~\ref{sec:FPLossMechanism}). The distillation loss function enforces the student model to learn soft targets of the teacher model~\cite{hinton2015distilling}. We have validated in Table~\ref{tb:vsSelfKD} that replacing the distillation loss function with our focus-picking loss function can further improve the performance of the state-of-the-art distillation methods.
\end{enumerate}

\section{Experiments}\label{experiments}

\subsection{Datasets}\label{datasets}

\begin{table}[!htpb]
	\renewcommand{\arraystretch}{1.2}
	\centering
	\caption{The datasets we used in this work. ``$ N $'' is the number of classes, ``\# Training'' and ``\# Validation'' represent the number of images in the training and validation sets, respectively.}\label{tb:dtst}
	\resizebox{0.7\textwidth}{!}{
		\begin{tabular}{l l r r}
			\hline\hline
			Datasets&{\textit{N}}&{\# Training}&{\# Validation}\\
			\hline
			1. MNIST~\cite{lecun1998gradient}		&10		&60,000						&10,000\\
			2. CIFAR-10~\cite{krizhevsky2009learning}	&10		&50,000						&10,000\\
			3. Imagewoof~\cite{imagewoof}	&10		&9,025						&3,929\\
			4. CIFAR-100~\cite{krizhevsky2009learning}	&100	&50,000						&10,000\\
			5. ImageNet Dogs~\cite{imgnetdog2016}	&118	&147,873	&5,900\\
			6. Stanford Dogs~\cite{khosla2011novel}	&120	&12,000	&8,580\\
			7. CUB-200~\cite{wah2011caltech}		&200	&5,994	&5,794\\
			8. Tiny ImageNet~\cite{le2015tiny}	&200	&100,000	&10,000\\
			9. ImageNet~\cite{russakovsky2015imagenet}	&1,000	&1,281,167	&50,000\\
			\hline\hline
		\end{tabular}
	}
\end{table}

Table~\ref{tb:dtst} lists the datasets used in our experiments, including both fine-grained datasets (Nos. 6 and 7) and datasets of different scales. We choose the fine-grained dataset considering that it contains a large number of confusing classes.

MNIST~\cite{lecun1998gradient} is a handwritten digits dataset with images of size 28$\times$28. It consists of 10 classes, with 60,000 training images and 10,000 validation images. 

CIFAR-10 and CIFAR-100~\cite{krizhevsky2009learning} have 10 and 100 classes, respectively. They both contain 50,000 training images and 10,000 validation images. The size of the images is 32$\times$32$\times$3.

Imagewoof~\cite{imagewoof} contains 10 dog breeds that are \textit{not} easy to classify. It contains 9,025 training images and 3,929 validation images.

ImageNet Dogs~\cite{imgnetdog2016} is composed of all dog breeds in the ImageNet dataset~\cite{russakovsky2015imagenet}. It has 118 classes with 147,873 images for training and 5,900 images for validation. All images have been down-sampled to 64$\times$64$\times$3 pixels.

Stanford Dogs~\cite{khosla2011novel} is a fine-grained dog breeds dataset. It contains 120 classes and 20,580 images, with 12,000 images for training and 8,580 ones for validation.

CUB-200~\cite{wah2011caltech} is a fine-grained bird species dataset. It contains 200 classes with 5,994 training images and 5,794 validation images.

Tiny ImageNet~\cite{le2015tiny} is a subset of the ImageNet dataset~\cite{russakovsky2015imagenet}. This dataset contains 200 classes, each of which has 500 training images and 50 validation images. All images have been down-sampled to 64$\times$64$\times$3 pixels.

ImageNet~\cite{russakovsky2015imagenet} is a large-scale dataset. It has 1,000 classes, 1.2 million training images and 50,000 validation images.

\subsection{Implementation Details}\label{sec:objTrainDetails}
We use LeNet-5~\cite{lecun1998gradient} for experiments on MNIST, use the standard ResNet-18~\cite{he2016deep} and MobileNetV2~\cite{sandler2018mobilenetv2} for experiments on Imagewoof, Stanford Dogs and CUB-200, and use the standard ResNet-18 and ResNet-34 for experiments on ImageNet. Because the standard ResNet-18 and MobileNetV2 down-sample input images by a factor of 32. To adapt them to small image sizes, we modify the first convolutional layer of ResNet-18 by setting the kernel size to 3$\times$3, the stride to 1, and the padding size to 1 as well. The max-pooling layer is removed. For MobileNetV2, we adjust the stride of the first and fourth convolutional layers to 1. As a result, the modified ResNet-18 and MobileNetV2 down-sample input images by a factor of 8. Therefore, we use the modified ResNet-18 and MobileNetV2 for experiments on CIFAR-10, CIFAR-100, Tiny ImageNet, and ImageNet Dogs. We follow the standard operation in MobileNetV2~\cite{sandler2018mobilenetv2} and uniformly multiply the width by 0.5 at each layer, which leads to a smaller and faster MobileNetV2 (0.94 M Parameters and 0.02 G FLOPs) than ResNet-18 (11.27 M Parameters and 4.46 G FLOPs). We use the identical network architecture for the baseline network and FocusNet if not otherwise specified.

For all experiments, we use stochastic gradient descent (SGD) with the initial learning rate of 0.1, the momentum of 0.9, and the weight decay of 0.0001. For MNIST, CIFAR-10, CIFAR-100, Imagewoof, Stanford Dogs, and CUB-200, we set the total number of epochs to 200 and divide the learning rate by 10 at epochs 100 and 150. For ImageNet Dogs, Tiny ImageNet, and ImageNet, we set the total number of epochs to 90 and divide the learning rate by 10 at epochs 30 and 60. We do not use data augmentation for MNIST, but standard augmentation for the others, \emph{i.e.} random cropping and flipping.

\subsection{Performance Evaluation}\label{sec:comparisons}
In the following, we evaluate the performance of FocusNet in terms of classification accuracy, compared with DropMax~\cite{lee2018dropmax} and a number of state-of-the-art knowledge distillation methods including \cite{yuan2020revisiting}, \cite{zhang2019your}, \cite{ji2021refine}, \cite{xu2019data}, \cite{yun2020regularizing}, \cite{lee2020self}, and \cite{lan2018knowledge}.

\subsubsection{Comparisons with DropMax}\label{sec:vsdropmax}
We validate our proposed FocusNet on all datasets and compare it with DropMax~\cite{lee2018dropmax}, because they both aim to focus on the confusing classes and ignore the irrelevant ones. For a fair comparison, we use the same network architecture for DropMax, the baseline network, and FocusNet.

\begin{table*}[h]
	\renewcommand{\arraystretch}{1.2}
	\centering
	\caption{Comparisons of top-1 validation accuracy (\%) with baseline and DropMax. The best ones are bolded, and the second-best ones are underlined. For MNIST, the network architecture is LeNet-5~\cite{lecun1998gradient}. For the other datasets, the network architecture is ResNet-18~\cite{he2016deep}. WOOF and INet.Dogs denote the Imagewoof and ImageNet Dogs datasets, respectively.}
	\label{tb:vsdropmax}
	\resizebox{\textwidth}{!}{
	\begin{tabular}{l|c|c|c|c|c|c|c|c}
		\hline \hline
		{Methods}	&MNIST	&CIFAR-10	&WOOF	&CIFAR-100	&CUB-200	&Stanford	&Tiny	&INet.Dogs                \\ 
		\hline		
		Baseline	&99.31	&92.43	&84.43	&74.52	&50.29	&63.32	&59.50	&70.76\\
		\hline
		DropMax~\cite{lee2018dropmax}&\underline{99.43}	&\underline{93.70}	&{\underline{86.87}}	&\underline{74.59}	&\underline{57.68}	&{\underline{64.02}}		&\underline{60.93}	&\underline{71.34}\\
		\hline
		FocusNet (Ours)&\textbf{99.56}	&\textbf{94.86}	&\textbf{86.94}	&\textbf{78.29}	&\textbf{63.43}	&\textbf{71.05}	&\textbf{64.49}	&\textbf{73.92}\\
		\hline\hline
	\end{tabular}
	}
\end{table*}

Table~\ref{tb:vsdropmax} shows the top-1 validation accuracy. We have two observations. (1) Both DropMax and our FocusNet improve the performances of the baseline network on all datasets, which implies that focusing on confusing classes is an effective strategy for classification. (2) Our FocusNet is more accurate than DropMax, especially on the fine-grained datasets.

{\begin{table*}[h]
\renewcommand{\arraystretch}{1.2}
\centering
\caption{Classification accuracy (\%) on ImageNet. The numbers in parentheses represent the accuracy difference between our FocusNet and the baseline network.}
\label{tb:imagenet}
\resizebox{0.75\textwidth}{!}{
\begin{tabular}{l|l|r|r}
	\hline \hline
	Model	&Methods	&Top-1	&Top-5\\
	\hline
	\multirow{2}{*}{ResNet-18}	&Baseline	&69.76	&89.08\\
	&FocusNet (Ours)&(+0.66) \textbf{70.42} & (+0.48) \textbf{89.56}\\
	\hline
	\multirow{2}{*}{ResNet-34}	&Baseline	&73.31	&91.42\\
	&FocusNet (Ours)&(+1.04) \textbf{74.35} &(+0.47) \textbf{91.89} \\
	\hline\hline		
\end{tabular}
}
\end{table*}
		
Table~\ref{tb:imagenet} shows the classification performance of our FocusNet on the ImageNet dataset. We did not compare it with DropMax because our experiments show that FocusNet constantly outperforms DropMax. Our FocusNet using ResNet-18 and ResNet-34 as network architectures consistently achieves better performance, which means that FocusNet scales well on different models.}

\subsubsection{Comparisons with Knowledge Distillation}
\label{sec:KD}
We compare our FocusNet with knowledge distillation~\cite{hinton2015distilling} because they both use knowledge from another network during training. However, as described in Section~\ref{sec:diffKD}, FocusNet is essentially different from the knowledge distillation. 

In our confusion-focusing mechanism, the architectures of the baseline branch and the FocusNet branch can be the same or different. Similarly, knowledge distillation can also use different network combinations. We also evaluate the classification performance in an extreme case that a poorly-trained and small model guides a large model.

\begin{table*}[h]
\renewcommand{\arraystretch}{1.2}
\centering
\caption{Model complexity comparison of different network architectures.}\label{tb:complexitybase}
\resizebox{0.65\textwidth}{!}{
\begin{tabular}{l|c|c|c}
	\hline\hline
	{Methods}	&{MobileNetV2}	&{ResNet-18}	&{ResNet-34}\\
	\hline
	{Params (M)}	&0.94	&11.27	&21.38\\
	\hline
	{FLOPs (G)}	&0.02	&4.46		&9.29\\
	\hline\hline
\end{tabular}}
\end{table*}

We select MobileNetV2~\cite{sandler2018mobilenetv2}, ResNet-18~\cite{he2016deep} or ResNet-34~\cite{he2016deep} as the network architecture in the following experiments. Table~\ref{tb:complexitybase} compares their model complexity. We choose three network combinations on most datasets, \emph{i.e.}, ``ResNet-18 $\rightarrow$ MobileNetV2'', ``ResNet-18 $\rightarrow$ ResNet-18'', and ``a poorly-trained MobileNetV2 $\rightarrow$ ResNet-18'', corresponding to standard, self and defective-reverse knowledge distillation, respectively. In addition, we also use ``ResNet-34 $\rightarrow$ ResNet-34'' for experiments on ImageNet to compare with the state-of-the-art self-knowledge distillation method.

\begin{table*}[h]
\renewcommand{\arraystretch}{1.2}
\centering
\caption{Comparisons of top-1 validation accuracy (\%) with standard knowledge distillation. The best ones are bolded, and the second-best ones are underlined. T and S denote teacher and student models, respectively.} \label{tb:vsStandardKD}
\resizebox{\textwidth}{!}{
	\begin{tabular}{l|c|c|c|c|c|c|c}
		\hline\hline
		Methods	&CIFAR-10	&WOOF	&CIFAR-100	&CUB-200	&Stanford	&Tiny	&INet.Dogs\\
		\hline
		T: ResNet-18&92.43	&84.43	&74.52	&50.29	&63.32	&59.50	&70.76\\
		\hline
		S: MobileNetV2&87.74	&81.99	&69.06	&52.71	&60.52	&52.77	&52.81\\
		\hline
		Standard KD	&\underline{88.20}	&\underline{82.76}	&\underline{71.44}	&\underline{58.03}	&\underline{64.73}	&\underline{56.60}	&\underline{64.20}\\
		\hline
		FocusNet (Ours)	&\textbf{90.45}	&\textbf{83.56}	&\textbf{74.49}	&\textbf{64.27}	&\textbf{66.71}	&\textbf{58.96}	&\textbf{65.78}\\
		\hline\hline	
	\end{tabular}
}
\end{table*}

We use ResNet-18 to guide MobileNetV2. Table~\ref{tb:vsStandardKD} compares the proposed FocusNet and standard knowledge distillation. Because we only evaluate the student model, comparisons among MobileNetV2, standard knowledge distillation and FocusNet are fair. Top-1 validation accuracy demonstrates that FocusNet is superior to standard knowledge distillation, especially on the fine-grained datasets, CUB-200, for example. 

\begin{table*}[h]
\renewcommand{\arraystretch}{1.2}
\centering
\caption{Model complexity comparisons with self-knowledge distillation methods. The network architecture is ResNet-18.}
\label{tb:complexityself}
\resizebox{\textwidth}{!}{
	\begin{tabular}{l|c|c|c|c|c|c|c|c|c}
		\hline\hline
		{Methods}	&{Baseline} &{DDGSD} &{CS-KD}	&{SLA-SD}&{Tf-KD}	&{ONE}		&{BYOT}		&{FRSKD}	&{FocusNet (Ours)}\\
		\hline
		{Params (M)}&{11.27}&{11.27}	&{11.27}	&{11.58}&{11.27}&{11.27}	&{12.49}	&{17.75}	&{11.27}	\\
		\hline
		{FLOPs (G)}	&{4.46}	&{4.46} &{4.46}	&{4.46}	&{4.46}&{4.46}&{4.68}&{7.62}	&{4.46}\\
		\hline\hline
	\end{tabular}	
}
\end{table*}

We adopt ResNet-18 to guide ResNet-18, and compare the proposed FocusNet with seven representative self-knowledge distillation methods including three data-augmentation-based methods (DDGSD~\cite{xu2019data}, CS-KD~\cite{yun2020regularizing}, SLA-SD~\cite{lee2020self}) and four methods using auxiliary networks (Tf-KD~\cite{yuan2020revisiting}, ONE~\cite{lan2018knowledge}, BYOT~\cite{zhang2019your}, FRSKD~\cite{ji2021refine}). Table~\ref{tb:complexityself} compares the model complexity between FocusNet and these self-distillation methods.

\begin{table}[h]
\renewcommand{\arraystretch}{1.2}
\centering
\caption{Comparisons of top-1 validation accuracy (\%) with seven self-distillation methods. The best ones are bolded, and the second-best ones are underlined. The network architecture is ResNet-18.}\label{tb:vsSelfKD}
\resizebox{\textwidth}{!}{
	\begin{tabular}{l|c|c|c|c|c}
		\hline\hline	
		{Methods}	&{CIFAR-100}	&{CUB-200}	&{Stanford Dogs}	&{Tiny}	&{ImageNet Dogs}\\
		\hline
		{Baseline~\cite{he2016deep}}	&{74.52}	&{50.29}	&{63.32}	&{59.50}	&{70.76}\\
		\hline
		{Tf-KD~\cite{yuan2020revisiting}}	&{76.86}	&{57.23}	&{66.55}	&{60.02}	&{71.64}\\
		\hline
		{ONE~\cite{lan2018knowledge}}	&{76.67}	&{54.71}	&{65.39}	&{62.33}	&{72.56}\\
		\hline
		{DDGSD~\cite{xu2019data}}	&{76.61}	&{58.49}	&{69.00}	&{61.59}	&{71.98}\\
		\hline
		{BYOT~\cite{zhang2019your}}	&{76.68}	&{58.66}	&{68.82}	&{63.10}	&{72.90}\\
		\hline
		{CS-KD~\cite{yun2020regularizing}}	&{77.19}	&{64.34}	&{68.91}	&{60.44}	&{72.76}\\
		\hline
		{SLA-SD~\cite{lee2020self}}	&{77.52}	&{56.17}	&{67.30}	&{60.81}	&{72.78}\\
		\hline
		{FRSKD$ \backslash $F~\cite{ji2021refine}}	&{77.64}	&{62.29}	&{69.48}	&{63.84}	&{73.53}\\
		\hline
		{FRSKD~\cite{ji2021refine}}	&{77.71}	&{\underline{65.39}}	&{70.77}	&{64.35}	&\underline{74.31}\\
		\hline
		{FocusNet (Ours)}	&{\underline{78.29}}	&{63.43}	&{\underline{71.05}}	&{\underline{64.49}}	&{73.92}\\
		\hline
		{OurLoss+FRSKD}	&{\textbf{78.45}}	&{\textbf{67.19}}	&{\textbf{71.49}}	&{\textbf{64.92}}	&{\textbf{74.69}}\\
		\hline\hline
	\end{tabular} 
}
\end{table}

Table~\ref{tb:vsSelfKD} compares the classification performance of FocusNet with self-knowledge distillation methods. Results show that our FocusNet outperforms state-of-the-art self-knowledge distillation methods on the CIFAR-100, Stanford Dogs, and Tiny ImageNet datasets. In addition, we notice that FRSKD~\cite{ji2021refine} uses an efficient learning paradigm that simultaneously trains the classifier network and an auxiliary self-teacher network, and passes feature maps to each other. We adopt this paradigm and replace its distillation loss function with our proposed focus-picking loss function, called OurLoss+FRSKD. The results show that OurLoss+FRSKD achieves the best performances on all datasets, validating the advantage of our focus-picking loss function.

\begin{table*}[h]
\renewcommand{\arraystretch}{1.2}
\centering
\caption{Comparisons of validation accuracy (\%) with FRSKD on ImageNet. The best ones are bolded, and the second-best ones are underlined.}
\label{tb:vsFRSKDImageNet}
\resizebox{0.6\textwidth}{!}{
	\begin{tabular}{l|l|r|r}
		\hline \hline
		Model	&Methods	&Top-1	&Top-5\\
		\hline
		\multirow{3}{*}{ResNet-18}	&Baseline	&69.76	&89.08\\
		&FRSKD&\underline{70.17}	&\textbf{89.78}\\
		&FocusNet (Ours)&\textbf{70.42} &\underline{89.56}\\
		\hline
		\multirow{3}{*}{ResNet-34}	&Baseline	&73.31	&91.42\\
		&FRSKD	&\underline{73.75}	&\textbf{92.11}\\
		&FocusNet (Ours)&\textbf{74.35} &\underline{91.89} \\
		\hline\hline		
	\end{tabular}
}
\end{table*}

Table~\ref{tb:vsFRSKDImageNet} compares our FocusNet with FRSKD on the ImageNet dataset. Results show that our FocusNet outperforms the state-of-the-art method in the top-1 accuracy.

\begin{table*}[h]
\renewcommand{\arraystretch}{1.2}
\centering
\caption{Comparisons of top-1 validation accuracy (\%) with defective-reverse knowledge distillation. The best ones are bolded, and the second-best ones are underlined. Pt-MobileNetV2 and ResNet-18 are the poorly-trained teacher and student models, respectively.} \label{tb:vsReverseKD}
\resizebox{\textwidth}{!}{
	\begin{tabular}{l|c|c|c|c|c|c|c}
		\hline\hline
		{Methods}	&{CIFAR-10}	&{WOOF}	&{CIFAR-100}	&{CUB-200}	&{Stanford}	&{Tiny}	&{INet.Dogs}\\
		\hline
		T: Pt-MobileNetV2&{46.93}	&{41.33}	&{35.65}	&{35.43}	&{37.98}	&{32.14}	&{33.29}\\
		\hline
		S: ResNet-18	&\underline{92.43}	&\underline{84.43}	&\underline{74.52}	&\underline{50.29}	&\underline{63.32}	&\underline{59.50}	&\underline{70.76}\\
		\hline
		De-reverse KD	&55.50	&51.69	&46.22	&42.89	&45.50	&38.45	&39.39\\
		\hline
		FocusNet (Ours)&\textbf{93.78}	&\textbf{86.13}	&\textbf{77.11}	&\textbf{62.34}	&\textbf{66.88}	&\textbf{62.65}	&\textbf{72.10}\\
		\hline\hline
	\end{tabular}
}
\end{table*}

We use the poorly-trained MobileNetV2 to guide ResNet-18 and compare the proposed FocusNet with the defective-reverse knowledge distillation. The poorly-trained MobileNetV2 means that we only train it a few epochs and do not decay the learning rate during training. Specifically, we train 1 epoch for CIFAR-10, 30 epochs for Imagewoof, 5 epochs for CIFAR-100, 50 epochs for CUB-200 and Stanford Dogs, and 10 epochs for Tiny ImageNet and ImageNet Dogs. Table~\ref{tb:vsReverseKD} shows that knowledge distillation is not suitable for this kind of network combination. However, FocusNet significantly improves the baseline ResNet-18. The differences in performance are mainly due to the difference in design motivations. As described in Section~\ref{sec:diffKD}, the knowledge distillation aims to train an easy-deployed student model to match the teacher's performance. However, FocusNet aims to focus on confusing classes. The credibility of confusing classes mainly depends on the top-\textit{k} accuracy instead of the top-1 accuracy, so the poorly-trained MobileNetV2 has less influence on our FocusNet than on the knowledge distillation.

\subsection{Impacts of the Inaccurate Baseline Network on FocusNet}
\label{sec:inaacuractBase}

\begin{figure}[t]
\centering
\includegraphics[width=\linewidth]{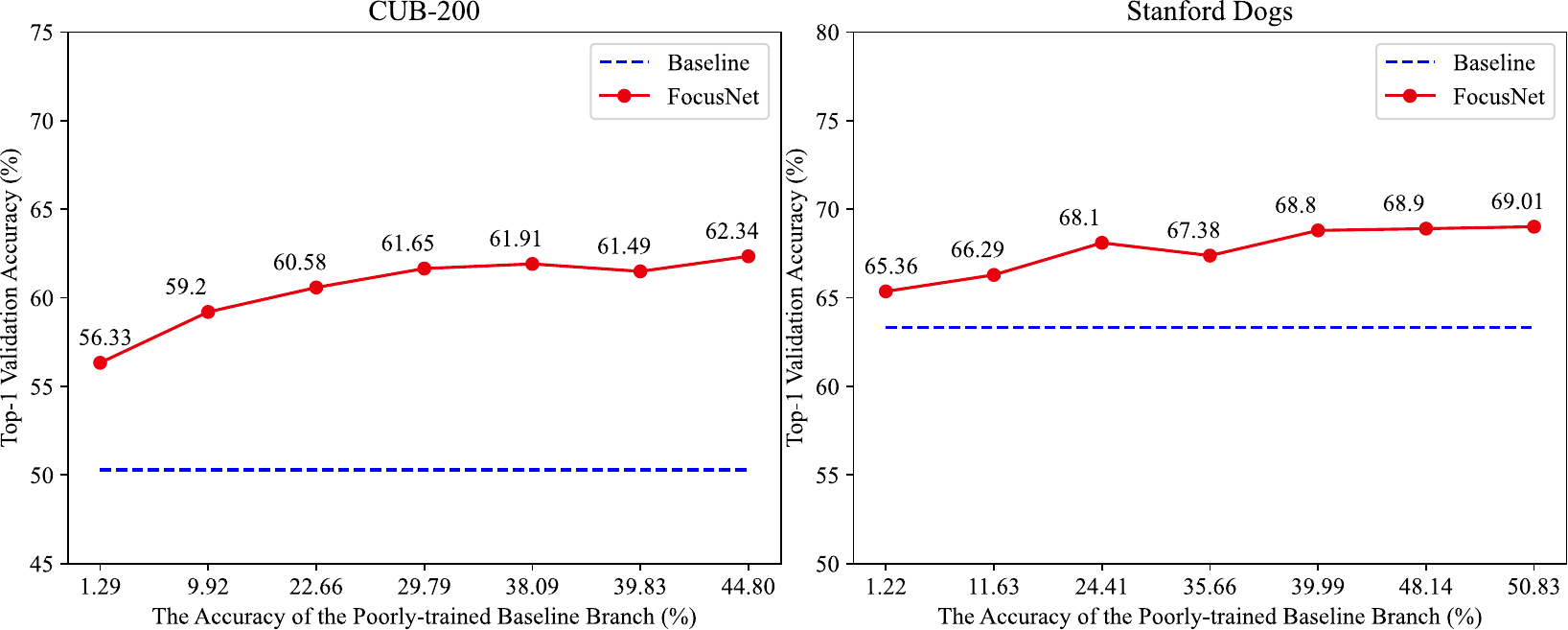}
\caption{The impact of inaccurate baseline branch on FocusNet. The poorly-trained baseline network denotes that we stop training at epochs 1, 10, 20, 30, 40, 50 and 60. The blue dash line and solid red line both indicate accuracy at epoch 200. The network architecture is ResNet-18. The benchmark datasets are CUB-200 and Stanford Dogs in (a) and (b), respectively.}\label{fig:inaacuractBase}
\end{figure}

We investigate the impact of confusing classes credibility on FocusNet by using a poorly-trained baseline network to guide FocusNet. In the following experiments, we select ResNet-18 as the architecture of both the baseline and FocusNet. The poorly-trained baseline ResNet-18 means that we only train it a few epochs and do not decay the learning rate during training. FocusNet denotes we train ResNet-18 with our focus-picking loss function. Fig.~\ref{fig:inaacuractBase} shows the impact of the inaccurate baseline network branch on FocusNet. The performance of FocusNet generally improves as the accuracy of the baseline network increases. Furthermore, even with a highly inaccurate pre-trained network (1.29\% for CUB-200 and 1.22\% for Stanford Dogs), FocusNet still outperforms the baseline (blue dashed line). We attribute this phenomenon to FocusNet, which emphasizes the contribution of difficult samples on the one hand, and focuses on the most confusing classes on the other. The credibility of confusing classes mainly depends on the top-\textit{k} accuracy rather than the top-1 accuracy.

\subsection{Ablation Study}
\begin{table}[h]
\renewcommand{\arraystretch}{1.2}
\centering
\caption{Ablation study results. The results are top-1 validation accuracy (\%). The network architecture is ResNet-18. The best performing strategies are in bold.}~\label{tb:ablation}
\resizebox{\textwidth}{!}{
\begin{tabular}{l|c|ccc|c|c}
	\hline\hline
	{Methods} &{Cross-entropy}	&{$\mathcal{L}_\mathrm{cls}$}	&{$\mathcal{R}_\mathrm{attention}$}	&{$\mathcal{R}_\mathrm{entropy}$}	&{CUB-200}	&{Stanford}\\
	\hline
	{Baseline}	&{$\checkmark$}	&{}	&{}	&{}	&{50.29}	&{63.32}\\
	\hline
	\multirow{3}{*}{Ablations}	&{}	&{$\checkmark$}	&{}	&{}	&{56.82}	&{64.07}\\
	{}			&{}	&{$\checkmark$}	&{}	&{$\checkmark$}	&{61.10}	&{68.76}\\
	{}	&{}	&{$\checkmark$}	&{$\checkmark$}	&{}	&{61.27}	&{67.21}\\
	\hline
	{FocusNet (Ours)}	&{}	&{$\checkmark$}	&{$\checkmark$}	&{$\checkmark$}	&{\textbf{63.43}}	&\textbf{71.05}\\
	\hline\hline
\end{tabular}
}
\end{table}
	
In this section, we investigate the necessity of each term of our proposed focus-picking loss function. Table~\ref{tb:ablation} shows the top-1 validation accuracy of ResNet-18 on the CUB-200 and Stanford Dogs datasets using different loss function combinations. $\mathcal{L}_\mathrm{cls}$ improves the baseline by emphasizing the importance of difficult samples. $\mathcal{R}_\mathrm{attention}$ and $\mathcal{R}_\mathrm{entropy}$ both contribute to better performances by encouraging FocusNet to focus on confusing classes and prevent over-fitting. Our FocusNet outperforms the baseline by a large margin, demonstrating the importance of each term of our focus-picking loss function.

\section{Conclusions}
In this work, we observe that a classification network trained with one-hot labels can pay attention to the target image region even if it makes an incorrect prediction. To enable the network to better discriminate between confusing classes, we propose a confusion-focusing mechanism and a focus-picking loss function by considering confusing inter-class interactions. The confusion-focusing mechanism is implemented in a two-branch network architecture, \emph{i.e.} a baseline branch and a FocusNet branch. The FocusNet branch does not restrict to specific network architectures and can adopt common models. Its main advantage is that it improves the performance of networks without increasing model complexity. Fine-grained classification and computational resource-limited scenes can benefit from our method because our method encourages networks to focus on the most confusing classes and does not impose additional deployment burdens at test time. In addition, we have demonstrated that FocusNet can still improve performance even with a highly-inaccurate pre-trained network. Experimental results have validated that our FocusNet performs well on a number of challenging datasets, and that the focus-picking loss function can further improve the state-of-the-art techniques. 

A limitation of our current confusion-focusing mechanism is its ineffective two-stage training strategy. In the future we will work towards an optimized one-stage scheme that can further improve training efficiency and classification accuracy.

%%%%%%%%%%%%%%%%%%%%%%%%%%%%%%%%%%%%%%%%%%%%%%%%%%%
%%% 				References					%%%

\bibliographystyle{elsarticle-num}
\bibliography{PR_REF}

\end{document}